\documentclass{AISB2008}
\usepackage{times}
\usepackage{graphicx}
\usepackage{latexsym}
\usepackage{balance}
\usepackage{adjustbox,lipsum}
\usepackage{rotating}

\begin{document}

\title{A platform-independent robot control architecture for multiple therapeutic scenarios}

\author{Hoang-Long Cao\institute{Vrije Universiteit Brussel, Robotics \& Multibody Mechanics Research Group, Pleinlaan 2, 1050 Brussels, Belgium}, Pablo G\'{o}mez Esteban\footnotemark[1], Albert De Beir\footnotemark[1], Greet Van de Perre\footnotemark[1],\\
	Ramona Simut\institute{Vrije Universiteit Brussel, Department of Clinical and Life Span Psychology Group, Pleinlaan 2, 1050 Brussels, Belgium} \and Bram Vanderborght\footnotemark[1]}

\maketitle

\begin{abstract}
While social robots are developed to provide assistance to users through social interactions, their behaviors are dominantly pre-programmed and remote-controlled. Despite the numerous robot control architectures being developed, very few offer reutilization opportunities in various therapeutic contexts. To bridge this gap, we propose a robot control architecture to be applied in different scenarios taking into account requirements from both therapeutic and robotic perspectives. As robot behaviors are kept at an abstract level and afterward mapped with the robot's morphology, the proposed architecture accommodates its applicability to a variety of social robot platforms.

\end{abstract}

\section{INTRODUCTION}
Social robots are developed to provide assistance to users through social interaction with appropriate behaviors, and expected to exhibit natural-appearing social manners and to enhance the quality of life for broad populations of users \cite{tapus2007socially}. Robots with a high degree of autonomy are particularly desirable in Robot-Assisted Therapy (RAT) \cite{thill2012robot}. Currently, robots' behaviors are dominantly pre-programmed and remote-controlled (Wizard-of-Oz technique) \cite{cabibihan2013robots,dautenhahn2007methodology,huijnen2016mapping,riek2015robotics}. In an attempt to address the shortcomings of these techniques, numerous architectures have been developed to aid social robots in autonomous decision making, e.g. \cite{belpaeme2012multimodal,Cao2014,feil2009toward,mead2010architecture, wada2008robot,wainer2014pilot}. Among these architectures, the behavior-based approach \cite{arkin1998behavior} is dominated while the robot behaviors should go further into adaptive behaviors to create a more efficient interaction e.g. natural language interaction, learning to generate more appropriate and personalized behaviors \cite{franccois2009using,thill2012robot}. Additionally, most of the efforts nevertheless employ ad-hoc solutions for particular therapies and robot platforms, and thus cannot be applied to any other therapeutic situations. To fill this gap, this paper proposes a platform-independent robot control architecture applicable in multiple scenarios based on requirements from therapeutic and robotic perspectives. We do not aim to improve the technical or therapeutic performance of the robot behavior, but to enable the robot control architecture to be easily applicable for different scenarios and robot platforms.

\section{STATE OF THE ART} 

Researchers have developed a number of architectures to aid social robots in autonomous decision making for various therapeutic applications with different roles e.g. companion, therapeutic play partner, coach. Companion robots are typically in the shape of animals, e.g. PARO, AIBO, with basic reactive behaviors and emotions to interact with children with autism \cite{stanton2008robotic}, and elderly with dementia \cite{wada2008robot} or loneliness \cite{banks2008animal}, etc. Some studies focus on using robots as therapeutic play partners mainly with children with autism in which the robots are used along with human therapists \cite{rabbitt2014integrating}. Architectures used in these studies are symbolic e.g. sense-plan-act (KASPAR) \cite{wainer2014pilot}, homeostatic-based with physiological and affectional needs (Probogotchi) \cite{Simut2016}, arbitration between task-based action and social/emotional action (Bandit and Pioneer) \cite{feil2008b,feil2009toward}; or neural networks e.g. Robota \cite{calinon2003pda,billard2007building}, FACE \cite{pioggia2007human}. Robots can be also used as coaches in therapies e.g. physical exercise (Bandit and Pioneer) \cite{fasola2013socially}, rehabilitation (Bandit and Pioneer) \cite{wade2011socially}, children affected by diabetes (NAO) \cite{belpaeme2012multimodal}. Robot behaviors in these applications are generated taking into account both the scenario and user's information such as user's performance or emotional feedback. Overall, most of the above-mentioned studies employ ad-hoc solutions for particular therapies and robot platforms, and thus cannot be applied to any other therapeutic situations.

\section{ARCHITECTURE REQUIREMENTS} \label{sec:sec_requirements}
\begin{figure}[t]
	\centerline{\includegraphics[width=\columnwidth]{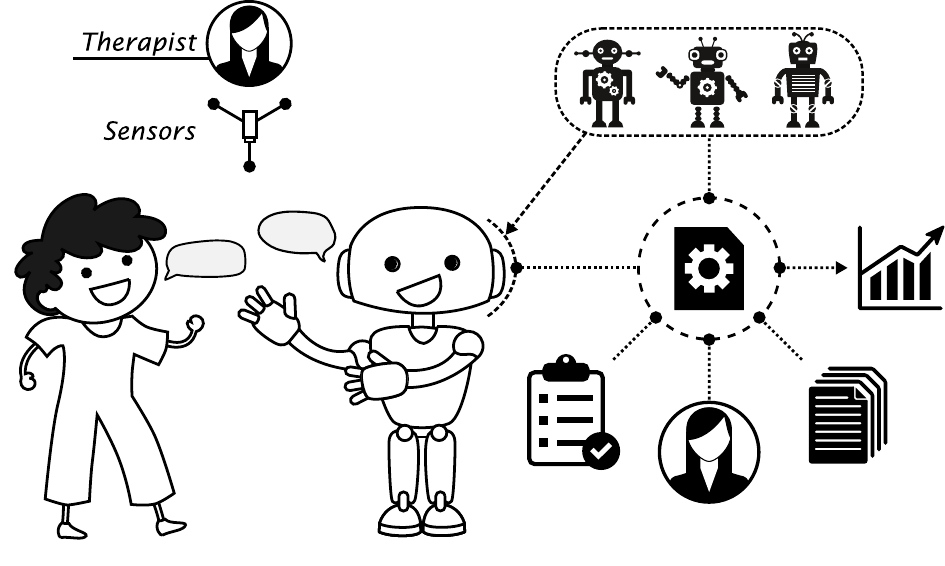}}
	\caption{Elements related to a platform-independent robot control architecture for multiple therapeutic scenarios: interaction protocol and goal, therapist, different scenarios, data for analysis, and different robot platforms.}
	\label{fig:fig_requirements}
\end{figure}

Robot-assisted therapy research requires interdisciplinary collaboration. A robot control architecture for RAT should meet requirements from both therapeutic and robotic perspectives. Okamura \textit{et. al} suggested ten desired system capabilities the robotic systems must have for ideal integration into medicine and healthcare focusing on technical issues \cite{okamura2010medical}. Therapists, on the other hand, pay more attention to the therapeutic process and ethical issues \cite{david2014robot,coeckelbergh2015survey}. We hereby summarize different perspectives into five main requirements for a robot control architecture, as a part of the robotic system, to enable the system to acquire the desired capabilities. Figure \ref{fig:fig_requirements} illustrates the elements related to these requirements.

\subsection{Sustaining user's motivation and engagement}
User's motivation and engagement significantly influence the therapeutic performance. Both extrinsic motivation (driven by external rewards) and intrinsic motivation (self-triggered based on enjoyment and satisfaction) need to be maintained during all phases of the therapeutic process e.g. diagnosis, intervention, prevention \cite{david2014robot,fasola2013socially}. From a therapeutic perspective, the human-robot interaction should be personal (e.g. personality adaptation, personal information, performance history) and have an interesting play scenario with different levels of difficulty \cite{fasola2013socially,pop2014enhancing,tapus2008user}. Taking these inputs, the robot behavior controller creates a fluid interaction with the user generating believable and coherent behaviors.

\subsection{Achieving the goal of interaction}
While most robot behaviors are generated based on the robot's ``well-being", robots used in RAT are designed to help a user obtain particular therapeutic goals with well-structured interaction protocols. The robot's decision making, therefore, has to follow the interaction protocol. Moreover, the robot behaviors should strictly conform to ethical issues (e.g. vocabulary, cognition and affect, decision making and action) defined in user-caregiver interaction manuals or by experts \cite{malle2015integrating}. To ensure the goal of interaction to be reached, the robot's operation should always be under the guidance of human therapists \cite{coeckelbergh2015survey}. 

\subsection{Therapeutic scenario-independence}
In order to go beyond ad-hoc solutions for particular therapies, the robot behavior architecture should be highly adaptable to different therapeutic scenarios. Architecture modules developed for a scenario might be reusable for another one without significant modification. 

\subsection{Platform-independence}
Being platform-independent is another requirement to move beyond ad-hoc solutions. A recent trend in social robotics focuses on coding robot behaviors by using parameters e.g. \cite{le2011design,salem2010generating,van2015development}.  Rather than controlling actuators specific to one robot platform, the architecture will prescribe parameters in descriptions and representations that are common across all platforms. Afterward, these robot non-specific commands will be translated into robot-specific actions. Consequently, behaviors developed for a certain robot can be transferred to another without reprogramming.

\subsection{Providing data for analysis}
In RAT research, therapists and roboticists might have different interests in extracted data. The robot behavior architecture should provide data (e.g. therapeutic performance, user's performance history, robot operation) recorded in structured forms to different parties.

\section{ARCHITECTURE DESIGN}

\subsection{Architecture's key specifications}
We propose five key specifications that a platform-independent robot control architecture should have to fulfill the requirements defined in Section \ref{sec:sec_requirements}:
\begin{enumerate}
	\item Human-robot personality adaptation
	\item User's profile and affect influencing behavior generation and realization
	\item Platform-independent behavior
	\item Supervised-autonomy	
	\item Modular multi-layer behavior architecture
\end{enumerate}

The first two specifications related to user's personality and emotion create a personal interaction and hence contribute to sustain user's motivation and engagement. The platform-independent behavior directly supports the system to be independent with the robot platform. Supervised-autonomy allows therapists to control the whole process of the robot system to ensure that the therapeutic goal can be reached. Lastly, designing an architecture with a modularity approach mainly contributes to therapeutic scenario-independence but more or less supports the other requirements.

\subsection{Architecture's working principle}

\begin{figure}
	\centerline{\includegraphics[width=\columnwidth]{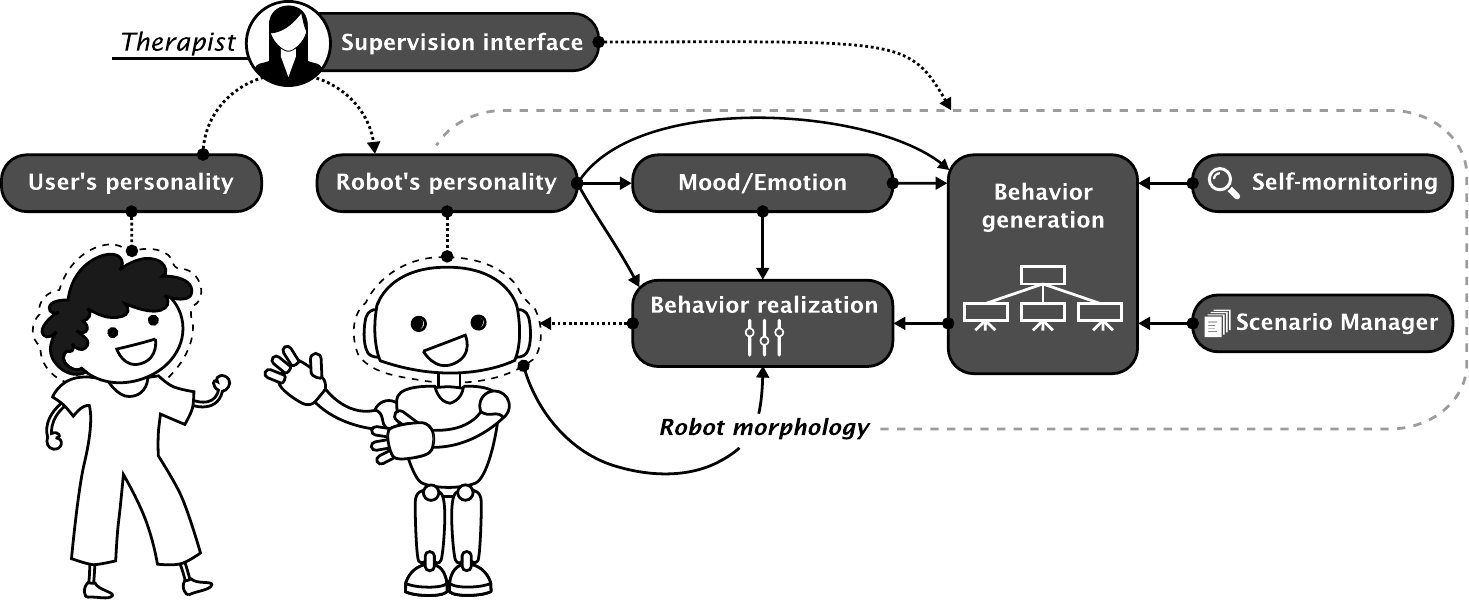}}
	\caption{Architecture's working principle. Robot behavior is generated taking into account user's profile, robot platform, scenario, physical and ethical limits. The whole system is supervised by therapist.}
	\label{fig:fig_working_principle}
\end{figure}

Figure \ref{fig:fig_working_principle} illustrates the working principle of the architecture to obtain the five aforementioned specifications. The architecture should be designed with a robot's personality module in which its parameters are adapted to user's personality. The personality determines the varied rates of mood and emotion when a certain event occurs. The behavior is generated based on the robot's personality and the user profile, therapeutic scenario, and self-monitoring rules. The generated behavior is realized taking into account the personality and robot's morphology. The whole system is supervised by a therapist.

\subsection{Design of a platform-independent robot control architecture for multiple therapeutic scenarios}

\begin{figure*}
	\centerline{\includegraphics[width=0.88\textwidth]{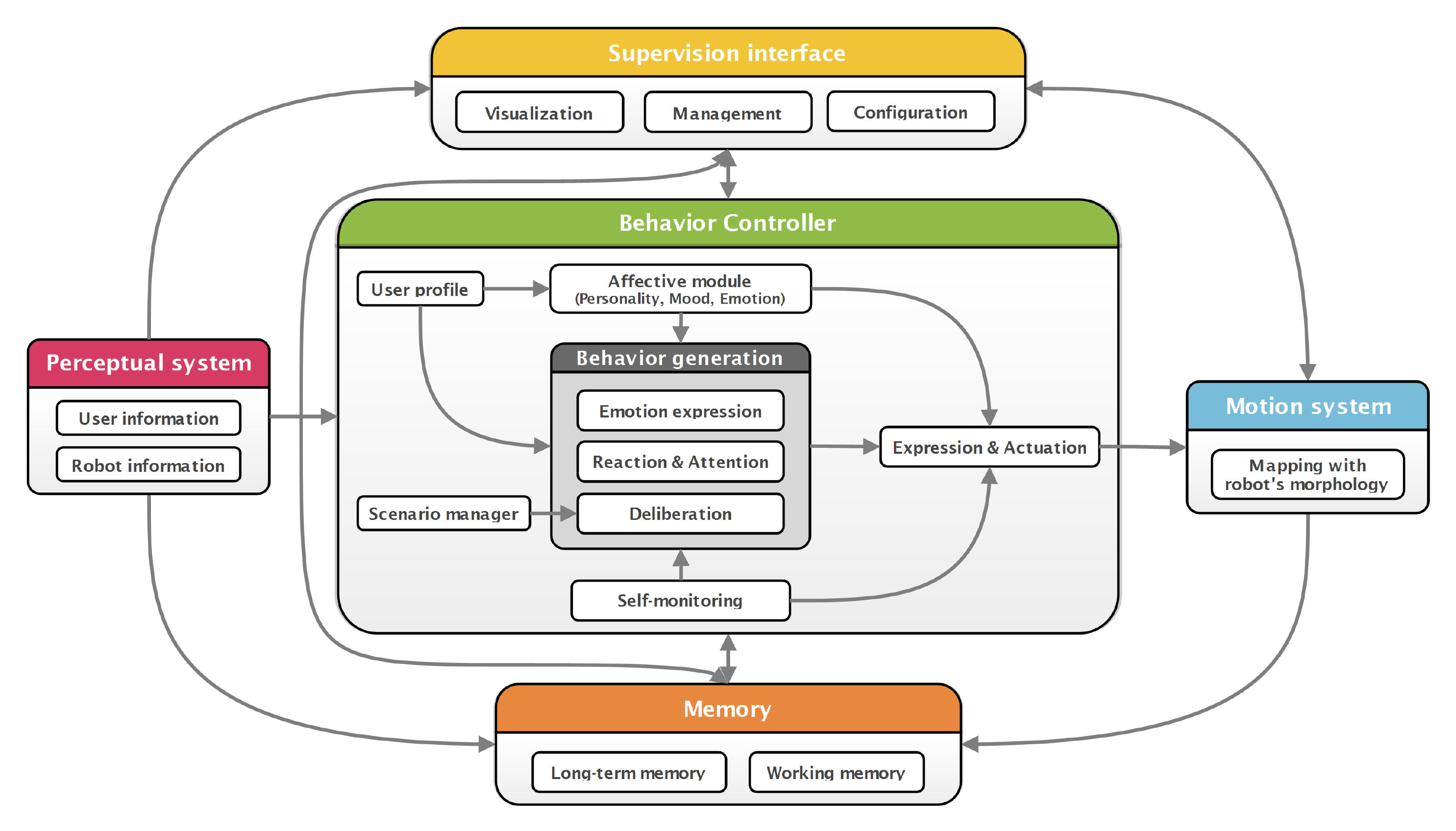}}
	\caption{Description of the robot control architecture. Depending on the therapy and robot platform, therapeutic supervisors decide which modules to use and the processing mechanisms for each module.}
	\label{fig:fig_architecture}
\end{figure*}

Following the working principle, we design an architecture with modules as depicted in Figure \ref{fig:fig_architecture}. The behavior controller, the center of the architecture, takes inputs from the perceptual system, user profile, scenario manager, supervision interface, and the memory - to generate proper behaviors.

The \textit{Perceptual system} is responsible for gathering information concerning the human-robot interaction. The system receives raw data from sensors (e.g. camera, touch sensors) and then interprets into interaction events and user's performance (e.g. level of engagement) as inputs for the \textit{Behavior controller}.

The \textit{Behavior controller} includes a number of modules to generate robot behaviors. The behavior generation is the main module organized with different layers of behavior. The Emotion expression module displays the internal robot emotion managed by the Affective module. The Reactive and Attention module generates life-like behaviors, perceptual attention, and attention emulation to create the illusion of the robot being alive \cite{lazzeri2013towards,saldien2014motion}. The Deliberation module generates deliberative behavior to follow the therapeutic scenario if users are engaged in a task and otherwise to (re)engage them to the task. The robot behavior is therefore more adaptive to the user and the interaction context \cite{franccois2009using,thill2012robot}. The behavior generation in these three modules is checked by applying the technical and ethical rules in the Self-monitoring module. Later, the Expression and actuation module combines the three generated behaviors into a unified one. The unified behavior is abstract and coded by the Facial Action Coding System and Body Action Units \cite{van2015development}. Parameters of the unified behavior are adjusted taking into account influences from the affective module (personality, mood, and emotion). Finally the abstract behavior is translated into robot-specific actions in the \textit{Motion system}.

The architecture operation is visualized, managed and configured through the \textit{Supervision interface}. This will enable the therapist an ability to select the scenario, supervise the behaviors of the robot, and interrupt the robot's operation if necessary.

In this architecture, we do not specify which computational model is used in each module e.g. affect model, behavior generation mechanism, behavior mapping method. Depending on the therapy, therapeutic supervisors decide which modules to use and the processing mechanisms for each module.

\section{DISCUSSION AND CONCLUSION}

\begin{figure}
	\centering
	\graphicspath{{}}
	\noindent\adjustbox{max width=\columnwidth}{%
		\begin{tabular}{cccc}
			& \textbf{\Large{}ASIMO} & \textbf{\Large{}Justin} & \textbf{\Large{}NAO}\tabularnewline
			\begin{turn}{90}
				\textbf{\Large{}Happiness}
			\end{turn} & \includegraphics{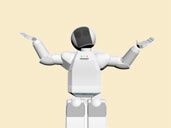} & \includegraphics{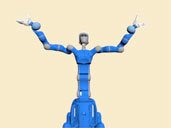} & \includegraphics{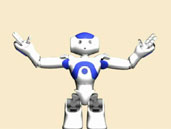}\tabularnewline
			\begin{turn}{90}
				\textbf{\quad\Large{}Sadness}
			\end{turn} & \includegraphics{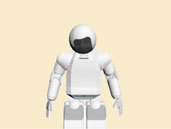} & \includegraphics{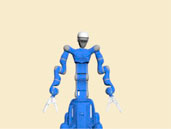} & \includegraphics{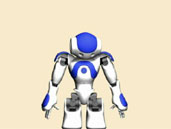}\tabularnewline
		\end{tabular}
	}
	\caption{Examples of upper-body emotional expressions in different robots: Happiness and Sadness \cite{van2015development}.}
	\label{fig:fig_gestures}
	
	\centering
	\includegraphics[height = 3.6cm]{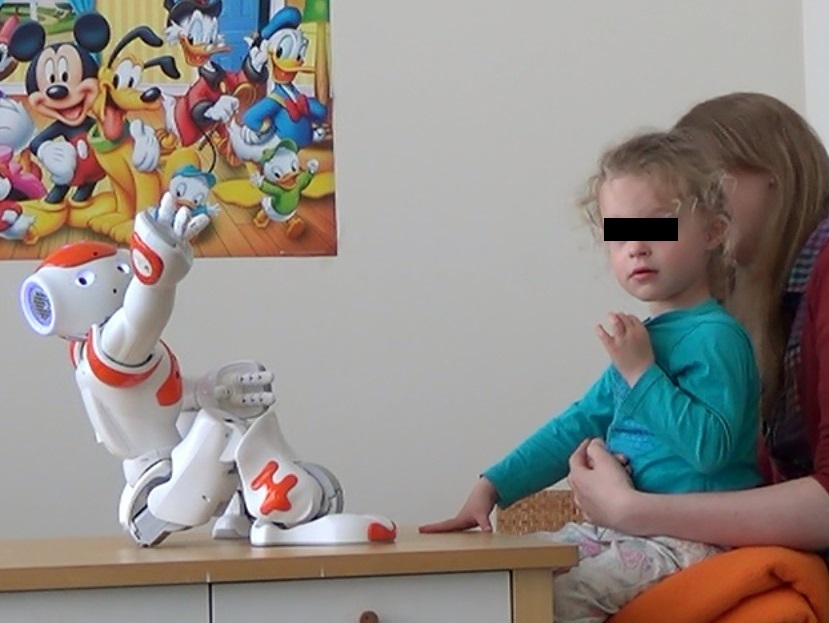}
	\includegraphics[height = 3.6cm]{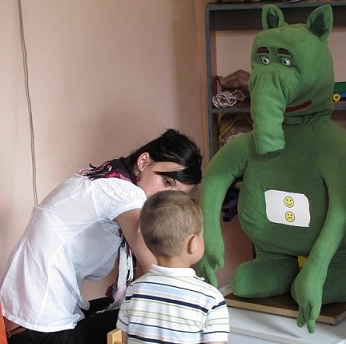}
	\caption{Child-robot interaction with NAO (left) and Probo (right).}
	\label{fig:fig_robots}
\end{figure}

This paper proposes the design of a platform-independent robot control architecture for multiple therapeutic scenarios. Requirements from literature of therapy and robotics have been analyzed in order to decide appropriate modules for the architecture. The selection of computational models and processing mechanisms for modules are dependent on the therapeutic supervisors. Computational models for each module will be selected based on popular modules in human-robot interaction and virtual reality studies e.g. Five-Factor model of personality, circumplex model of affect, homeostatic drive theory for behavior decision. The integration of the modules in real-life application will face a number of technical challenges e.g. real-time communication and computation, the combination of action outputs from different modules. Other challenges might include correctly evaluating user's personality and solving conflict among different ethical rules, etc.

As for validation, the architecture will be implemented in robot-assisted autism therapy scenarios e.g. joint attention, turn taking, and imitation. The utilized robot platforms will be Nao and Probo \cite{vanderborght2012using} as depicted in Figure \ref{fig:fig_robots}. Based on our previous work, a number of modules will be re-utilized. The behavior generation will be based on the homeostatic drive approach \cite{Cao2014,Simut2016}. The generic method to generate actions for different robots in \cite{van2015development} will be applied in therapeutic contexts (Figure~\ref{fig:fig_gestures}). For this method to work in real-time is inevitably one of the biggest challenges foreseen. Experimental results will be used to evaluate the architecture performance e.g. comparing with the Wizard-of-Oz setup, assessing the applicability with different robot platforms and therapeutic scenarios.

\ack
The work leading to these results has received funding from the European Commission 7th Framework Program as a part of the project DREAM grant no. 611391, Fonds Wetenschappelijk Onderzoek, and Innovatie door Wetenschap en Technologie.

\bibliographystyle{abbrv}
\bibliography{aisb}

\begin{thebibliography}{10}

\bibitem{arkin1998behavior}
R.~C. Arkin.
\newblock {\em Behavior-based robotics}.
\newblock MIT press, 1998.

\bibitem{banks2008animal}
M.~R. Banks, L.~M. Willoughby, and W.~A. Banks.
\newblock Animal-assisted therapy and loneliness in nursing homes: use of
  robotic versus living dogs.
\newblock {\em Journal of the American Medical Directors Association},
  9(3):173--177, 2008.

\bibitem{belpaeme2012multimodal}
T.~Belpaeme, P.~E. Baxter, R.~Read, R.~Wood, H.~Cuay{\'a}huitl, B.~Kiefer,
  S.~Racioppa, I.~Kruijff-Korbayov{\'a}, G.~Athanasopoulos, V.~Enescu, et~al.
\newblock Multimodal child-robot interaction: Building social bonds.
\newblock {\em Journal of Human-Robot Interaction}, 1(2):33--53, 2012.

\bibitem{billard2007building}
A.~Billard, B.~Robins, J.~Nadel, and K.~Dautenhahn.
\newblock Building robota, a mini-humanoid robot for the rehabilitation of
  children with autism.
\newblock {\em Assistive Technology}, 19(1):37--49, 2007.

\bibitem{cabibihan2013robots}
J.-J. Cabibihan, H.~Javed, M.~Ang~Jr, and S.~M. Aljunied.
\newblock Why robots? a survey on the roles and benefits of social robots in
  the therapy of children with autism.
\newblock {\em International journal of social robotics}, 5(4):593--618, 2013.

\bibitem{calinon2003pda}
S.~Calinon and A.~Billard.
\newblock Pda interface for humanoid robots.
\newblock In {\em Proceedings of the IEEE International Conference on Humanoid
  Robots (Humanoids)}, number LSA3-CONF-2003-002, 2003.

\bibitem{Cao2014}
H.-L. Cao, P.~Gomez~Esteban, A.~De~Beir, R.~Simut, G.~Van~de Perre, D.~Lefeber,
  and B.~Vanderborght.
\newblock {ROBEE}: A homeostatic-based social behavior controller for robots in
  human-robot interaction experiments.
\newblock In {\em Robotics and Biomimetics (ROBIO), 2014 IEEE International
  Conference on}, pages 516--521, Dec 2014.

\bibitem{coeckelbergh2015survey}
M.~Coeckelbergh, C.~Pop, R.~Simut, A.~Peca, S.~Pintea, D.~David, and
  B.~Vanderborght.
\newblock A survey of expectations about the role of robots in robot-assisted
  therapy for children with asd: Ethical acceptability, trust, sociability,
  appearance, and attachment.
\newblock {\em Science and engineering ethics}, pages 1--19, 2015.

\bibitem{dautenhahn2007methodology}
K.~Dautenhahn.
\newblock Methodology and themes of human-robot interaction: a growing research
  field.
\newblock {\em International Journal of Advanced Robotic Systems}, 2007.

\bibitem{david2014robot}
D.~David, S.-A. Matu, and O.~A. David.
\newblock Robot-based psychotherapy: Concepts development, state of the art,
  and new directions.
\newblock {\em International Journal of Cognitive Therapy}, 7(2):192--210,
  2014.

\bibitem{fasola2013socially}
J.~Fasola and M.~Mataric.
\newblock A socially assistive robot exercise coach for the elderly.
\newblock {\em Journal of Human-Robot Interaction}, 2(2):3--32, 2013.

\bibitem{feil2008b}
D.~Feil-Seifer and M.~J. Mataric.
\newblock B3ia: A control architecture for autonomous robot-assisted behavior
  intervention for children with autism spectrum disorders.
\newblock In {\em Robot and Human Interactive Communication, 2008. RO-MAN 2008.
  The 17th IEEE International Symposium on}, pages 328--333. IEEE, 2008.

\bibitem{feil2009toward}
D.~Feil-Seifer and M.~J. Matari{\'c}.
\newblock Toward socially assistive robotics for augmenting interventions for
  children with autism spectrum disorders.
\newblock In {\em Experimental robotics}, pages 201--210. Springer, 2009.

\bibitem{franccois2009using}
D.~Fran{\c{c}}ois, K.~Dautenhahn, and D.~Polani.
\newblock Using real-time recognition of human-robot interaction styles for
  creating adaptive robot behaviour in robot-assisted play.
\newblock In {\em Artificial Life, 2009. ALife'09. IEEE Symposium on}, pages
  45--52. IEEE, 2009.

\bibitem{huijnen2016mapping}
C.~A. Huijnen, M.~A. Lexis, R.~Jansens, and L.~P. de~Witte.
\newblock Mapping robots to therapy and educational objectives for children
  with autism spectrum disorder.
\newblock {\em Journal of autism and developmental disorders}, pages 1--15,
  2016.

\bibitem{lazzeri2013towards}
N.~Lazzeri, D.~Mazzei, A.~Zaraki, and D.~De~Rossi.
\newblock Towards a believable social robot.
\newblock In {\em Biomimetic and Biohybrid Systems}, pages 393--395. Springer,
  2013.

\bibitem{le2011design}
Q.~A. Le, S.~Hanoune, and C.~Pelachaud.
\newblock Design and implementation of an expressive gesture model for a
  humanoid robot.
\newblock In {\em Humanoid Robots (Humanoids), 2011 11th IEEE-RAS International
  Conference on}, pages 134--140. IEEE, 2011.

\bibitem{malle2015integrating}
B.~F. Malle.
\newblock Integrating robot ethics and machine morality: the study and design
  of moral competence in robots.
\newblock {\em Ethics and Information Technology}, pages 1--14, 2015.

\bibitem{mead2010architecture}
R.~Mead, E.~Wade, P.~Johnson, A.~S. Clair, S.~Chen, and M.~J. Mataric.
\newblock An architecture for rehabilitation task practice in socially
  assistive human-robot interaction.
\newblock In {\em RO-MAN, 2010 IEEE}, pages 404--409. IEEE, 2010.

\bibitem{okamura2010medical}
A.~M. Okamura, M.~J. Mataric, and H.~I. Christensen.
\newblock Medical and health-care robotics.
\newblock {\em Robotics and Automation Magazine}, 17(3):26--27, 2010.

\bibitem{pioggia2007human}
G.~Pioggia, M.~Sica, M.~Ferro, R.~Igliozzi, F.~Muratori, A.~Ahluwalia, and
  D.~D. Rossi.
\newblock Human-robot interaction in autism: Face, an android-based social
  therapy.
\newblock In {\em Robot and Human interactive Communication, 2007. RO-MAN 2007.
  The 16th IEEE International Symposium on}, pages 605--612. IEEE, 2007.

\bibitem{pop2014enhancing}
C.~A. Pop, S.~Pintea, B.~Vanderborght, and D.~O. David.
\newblock Enhancing play skills, engagement and social skills in a play task in
  asd children by using robot-based interventions. a pilot study.
\newblock 2014.

\bibitem{rabbitt2014integrating}
S.~M. Rabbitt, A.~E. Kazdin, and B.~Scassellati.
\newblock Integrating socially assistive robotics into mental healthcare
  interventions: Applications and recommendations for expanded use.
\newblock {\em Clinical Psychology Review}, 2014.

\bibitem{riek2015robotics}
L.~D. Riek.
\newblock Robotics technology in mental health care.
\newblock {\em Artificial Intelligence in Behavioral and Mental Health Care},
  page 185, 2015.

\bibitem{saldien2014motion}
J.~Saldien, B.~Vanderborght, K.~Goris, M.~Van~Damme, and D.~Lefeber.
\newblock A motion system for social and animated robots.
\newblock {\em International Journal of Advanced Robotic Systems}, 11, 2014.

\bibitem{salem2010generating}
M.~Salem, S.~Kopp, I.~Wachsmuth, and F.~Joublin.
\newblock Generating multi-modal robot behavior based on a virtual agent
  framework.
\newblock In {\em Proceedings of the ICRA 2010 Workshop on Interactive
  Communication for Autonomous Intelligent Robots (ICAIR)}, 2010.

\bibitem{Simut2016}
R.~Simut, G.~Van~de Perre, J.~Saldien, C.~Pop, J.~Vanderfaeillie, D.~David,
  D.~Lefeber, and B.~Vanderborght.
\newblock Probogotchi: a novel edutainment device as a bridge for interaction
  between a child with asd and the typically developed sibling.
\newblock {\em Journal of Evidence-Based Psychotherapies}, 2016.
\newblock (accepted).

\bibitem{stanton2008robotic}
C.~M. Stanton, P.~H. Kahn~Jr, R.~L. Severson, J.~H. Ruckert, and B.~T. Gill.
\newblock Robotic animals might aid in the social development of children with
  autism.
\newblock In {\em Human-Robot Interaction (HRI), 2008 3rd ACM/IEEE
  International Conference on}, pages 271--278. IEEE, 2008.

\bibitem{tapus2007socially}
A.~Tapus, M.~J. Mataric, and B.~Scasselati.
\newblock Socially assistive robotics [grand challenges of robotics].
\newblock {\em Robotics \& Automation Magazine, IEEE}, 14(1):35--42, 2007.

\bibitem{tapus2008user}
A.~Tapus, C.~{\c{T}}{\u{a}}pu{\c{s}}, and M.~J. Matari{\'c}.
\newblock User--robot personality matching and assistive robot behavior
  adaptation for post-stroke rehabilitation therapy.
\newblock {\em Intelligent Service Robotics}, 1(2):169--183, 2008.

\bibitem{thill2012robot}
S.~Thill, C.~A. Pop, T.~Belpaeme, T.~Ziemke, and B.~Vanderborght.
\newblock Robot-assisted therapy for autism spectrum disorders with (partially)
  autonomous control: Challenges and outlook.
\newblock {\em Paladyn, Journal of Behavioral Robotics}, 3(4):209--217, 2012.

\bibitem{van2015development}
G.~{Van de Perre}, M.~{Van Damme}, D.~Lefeber, and B.~Vanderborght.
\newblock Development of a generic method to generate upper-body emotional
  expressions for different social robots.
\newblock {\em Advanced Robotics}, 29(9):597--609, 2015.

\bibitem{vanderborght2012using}
B.~Vanderborght, R.~Simut, J.~Saldien, C.~Pop, A.~S. Rusu, S.~Pintea,
  D.~Lefeber, and D.~O. David.
\newblock Using the social robot {Probo} as a social story telling agent for
  children with {ASD}.
\newblock {\em Interaction Studies}, 13(3):348--372, 2012.

\bibitem{wada2008robot}
K.~Wada, T.~Shibata, T.~Musha, and S.~Kimura.
\newblock Robot therapy for elders affected by dementia.
\newblock {\em Engineering in Medicine and Biology Magazine, IEEE},
  27(4):53--60, 2008.

\bibitem{wade2011socially}
E.~Wade, A.~Parnandi, R.~Mead, and M.~Matari{\'c}.
\newblock Socially assistive robotics for guiding motor task practice.
\newblock {\em Paladyn, Journal of Behavioral Robotics}, 2(4):218--227, 2011.

\bibitem{wainer2014pilot}
J.~Wainer, K.~Dautenhahn, B.~Robins, and F.~Amirabdollahian.
\newblock A pilot study with a novel setup for collaborative play of the
  humanoid robot kaspar with children with autism.
\newblock {\em International journal of social robotics}, 6(1):45--65, 2014.

\end{thebibliography}

\end{document}